\documentclass[letterpaper, 10 pt, conference]{ieeeconf}

\IEEEoverridecommandlockouts
\usepackage{cite}
\usepackage{amsmath,amssymb,amsfonts}
\usepackage{algorithm}
\usepackage{algorithmic}
\usepackage{graphicx}
\usepackage{textcomp}
\usepackage{xcolor}
\def\BibTeX{{\rm B\kern-.05em{\sc i\kern-.025em b}\kern-.08em
    T\kern-.1667em\lower.7ex\hbox{E}\kern-.125emX}}

\title{\LARGE \bfseries
TaCauchy: An Extensible FEM Framework for Vision-Based Tactile Simulation
}
\author{Hengfei Zhao$^{1}$\authorrefmark{1}, Yifan Xie$^{1}$\authorrefmark{1}, Junhao Gong$^{1}$, Yue Sun$^{2}$, Kai Zhu$^{2}$, Weihua He$^{2}$, \\ Shoujie Li$^{1}$, Haohuan Fu$^{1}$, Wenbo Ding$^{1}$\authorrefmark{2}%
\thanks{\authorrefmark{1}These authors contributed equally to this work.}%
\thanks{Accepted to IEEE/RSJ International Conference on Intelligent Robots and Systems (IROS), 2026.}%
\thanks{\authorrefmark{2}Corresponding author: Wenbo Ding (ding.wenbo@sz.tsinghua.edu.cn).}%
\thanks{$^{1}$Shenzhen International Graduate School, Tsinghua University, Shenzhen 518055, China.}%
\thanks{$^{2}$Huawei Inc., Shenzhen, China.}%
\thanks{\copyright~2026 IEEE. Personal use of this material is permitted. Permission from IEEE must be obtained for all other uses, in any current or future media, including reprinting/republishing this material for advertising or promotional purposes, creating new collective works, for resale or redistribution to servers or lists, or reuse of any copyrighted component of this work in other works.}%
}

\begin{document}

\maketitle
\thispagestyle{empty}
\pagestyle{empty}

\begin{abstract}
Vision-based tactile sensors require high-fidelity simulation for reinforcement learning, yet existing approaches struggle to provide accurate mechanical stress fields within GPU-accelerated robotics platforms. We present \textit{TaCauchy}, an extensible Finite Element Method (FEM) framework that integrates rigorous physics-based force computation into Isaac Sim. Built on the Unified Incremental Potential Contact (UIPC) solver, \textit{TaCauchy} directly computes Cauchy stress tensors from hyperelastic constitutive laws and projects them onto contact surfaces to obtain traction forces and pressure distributions—providing mechanical ground truth from first principles rather than empirical estimation. Our framework features automatic mesh generation with geometry-aware adaptive refinement and a modular sensor interface enabling rapid integration of diverse sensors (GelSight Mini, DIGIT, 9DTact) with minimal configuration. Performance benchmarks demonstrate 33.40 FPS for single environments and 555 FPS aggregate throughput across 60 parallel environments, with stress extraction overhead under 1 ms. Physical validation experiments show strong agreement between simulated and real tactile responses across force ranges from 1.2556 N to 4.7332 N, achieving SSIM above 0.93, confirming the framework's capability to provide accurate, physically-grounded force supervision for downstream robotic manipulation tasks.
\end{abstract}


\section{Introduction}\label{sec:introduction}

Vision-Based Tactile Sensors (VBTS) provide high-resolution local feedback essential for contact-rich manipulation \cite{yuan2017gelsight,lambeta2020digit,li2026biomimetic,li2024m3tac,li2025sandworm,li2026master,xie2026learning}. Deep reinforcement learning and sim-to-real transfer require high-fidelity simulation, as soft-body contact involves inherently non-linear, high-frequency interactions. While recent simulators have made significant progress \cite{nguyen2024tacex,du2024tacipc,chen2026univtac,li2025taccel}, a fundamental challenge remains: how to obtain accurate and reliable force information through rigorous mechanical computation while maintaining the computational efficiency required for large-scale parallel training, and simultaneously provide high-fidelity visual rendering of tactile sensor responses.

\begin{figure}[t]
\centering
\includegraphics[width=\columnwidth]{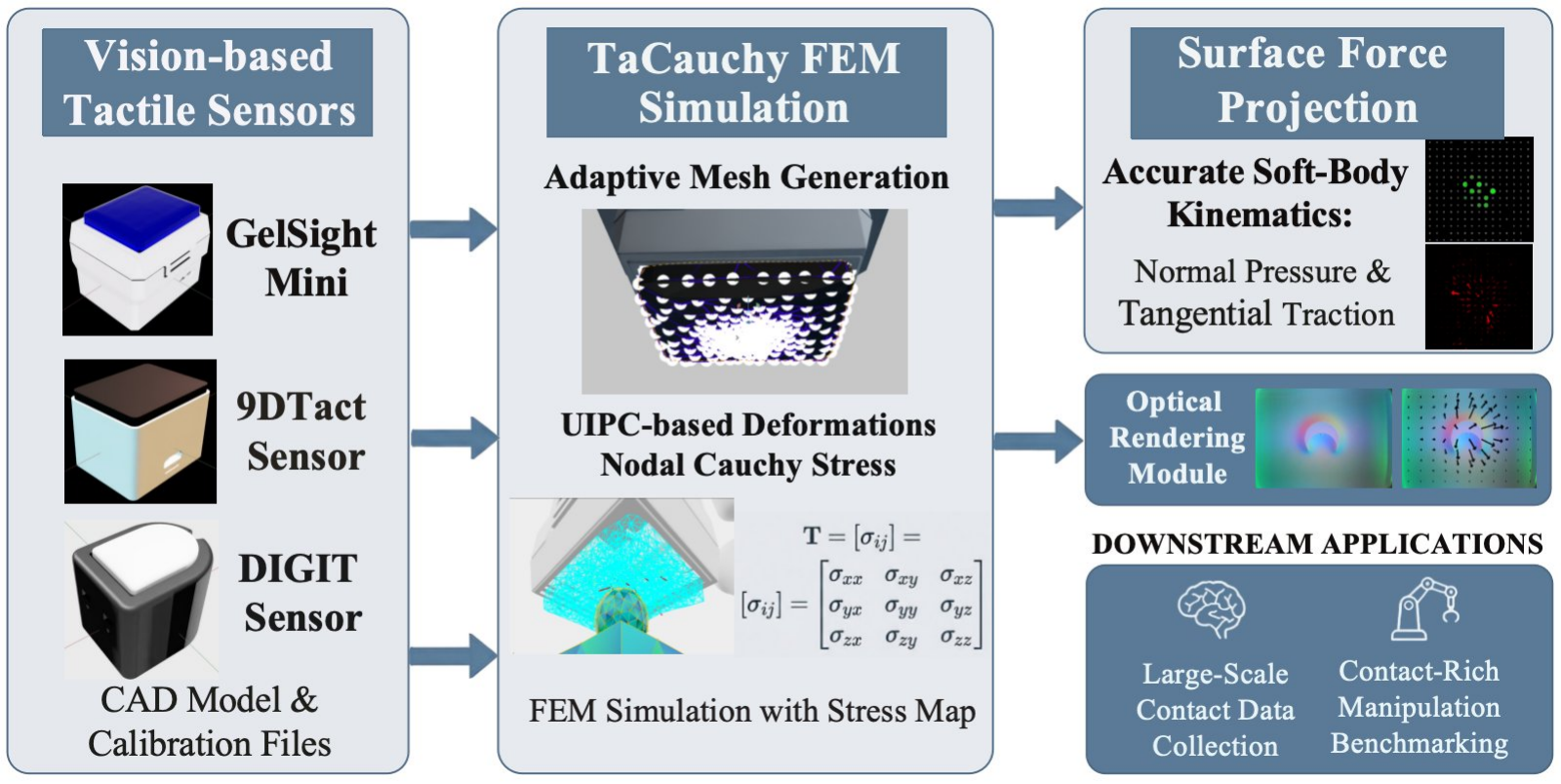}
\caption{TaCauchy: An extensible FEM framework integrating physics-based force computation into Isaac Sim. Our system combines UIPC soft-body dynamics, automatic adaptive mesh refinement, direct stress extraction from hyperelastic constitutive laws, and modular sensor interfaces, enabling efficient simulation of diverse tactile sensors (GelSight Mini, DIGIT, 9DTact) with accurate mechanical ground truth for reinforcement learning.}
\label{fig:overview}
\end{figure}

Existing approaches face trade-offs between physical accuracy and computational requirements. Some simulators operate outside the Isaac ecosystem \cite{du2024tacipc,si2024difftactile}, limiting their utility for large-scale RL training. Others sacrifice physical accuracy by simplifying contact models or lacking comprehensive force extraction capabilities \cite{akinola2025tacsl}. Additionally, purely optical-based simulators \cite{wang2022tacto,si2022taxim,zhao2024fots} cannot provide force information at all, leading to poor performance in contact-rich manipulation tasks that require accurate force feedback. The core technical challenge is designing a system that provides trustworthy mechanical ground truth through rigorous finite element analysis---including stress tensors, pressure distributions, and traction forces---while generating photorealistic tactile images and maintaining computational efficiency suitable for GPU-accelerated parallel training in Isaac Sim.

We introduce \textit{TaCauchy}, an extensible FEM framework that addresses these challenges by integrating physics-based force computation into the Isaac Sim ecosystem. Named after the fundamental Cauchy stress tensor \cite{gurtin1981introduction}, our framework directly calculates the true stress state from the deformation gradient field via hyperelastic constitutive laws. By projecting these tensors onto contact surfaces to obtain traction forces, \textit{TaCauchy} operates from first principles, providing trustworthy mechanical ground truth without the uncertainty of empirical estimations. Our framework features automatic mesh generation with geometry-aware adaptive refinement that concentrates computational resources at contact regions, a modular sensor interface enabling rapid integration of new sensors, and optional optical rendering for visual reference. We demonstrate extensibility by integrating three diverse sensors with minimal configuration changes.


Our contributions are summarized as follows:

\begin{itemize}
\item \textbf{TaCauchy Framework}: An extensible FEM framework integrating complete physics-based force computation into Isaac Sim, featuring direct calculation of Cauchy stress tensors, pressure, and traction forces from hyperelastic constitutive laws.
\item \textbf{Automated Mesh and Modular Design}: Automatic mesh generation with geometry-aware adaptive refinement and a modular sensor interface, enabling efficient integration of diverse tactile sensors with minimal configuration.
\item \textbf{Ecosystem Integration}: Demonstration of the framework's capability to support multiple sensor types (GelSight Mini, DIGIT, 9DTact) within the Isaac Lab ecosystem.
\end{itemize}

\section{Related Work}\label{sec:related_work}

\subsection{Tactile Simulation Approaches}
High-fidelity simulation is essential for scaling up contact-rich robotic manipulation. Early visuo-tactile simulators, such as TACTO \cite{wang2022tacto} and Taxim \cite{si2022taxim}, primarily relied on rigid-body penetration depth approximations or instance-based geometric mappings to generate tactile images. While computationally efficient, these optical rendering methods focus on producing visually realistic images but cannot compute physical quantities such as surface pressure, frictional traction forces, or the actual deformation of the silicone gel.

To achieve higher physical fidelity, recent research has shifted towards the Finite Element Method (FEM). However, different approaches obtain force information through fundamentally different mechanisms. Learning-based inverse methods such as iFEM \cite{ma2019dense} estimate force distributions by inverting observed gel deformations through trained models, introducing uncertainty and requiring extensive real-world calibration data. Simplified contact models like TacSL's Kelvin-Voigt formulation \cite{akinola2025tacsl} achieve over 200× speedup by replacing full FEM with spring-damper systems, trading physical accuracy for computational efficiency through empirical approximations. In contrast, our approach directly computes forces from first principles through hyperelastic constitutive laws, providing trustworthy mechanical ground truth without estimation uncertainty or empirical approximation.

\subsection{FEM-Based Tactile Simulators}
Several recent works have introduced FEM-based tactile simulation. TacIPC \cite{du2024tacipc} introduced inversion-free FEM solvers focusing primarily on image quality and marker displacement prediction. DIFFTACTILE \cite{si2024difftactile} emphasized differentiability for gradient-based optimization but operates as a standalone framework outside the Isaac ecosystem, limiting its utility for large-scale RL training. 

High-performance GPU-based platforms have emerged to support large-scale Reinforcement Learning. Taccel \cite{li2025taccel} integrates Incremental Potential Contact (IPC) and Affine Body Dynamics (ABD) to achieve massive parallelization with over 4000 parallel environments. TacEx \cite{nguyen2024tacex} pioneered the integration of soft-body simulation (GIPC) with optical rendering in Isaac Sim, providing a foundation for combining FEM mechanics with GPU-accelerated robotics platforms. UniVTAC \cite{chen2026univtac} offers a unified generation platform for downstream tasks with support for multiple sensors within the Isaac ecosystem.

However, these approaches face limitations in force computation completeness or ecosystem integration. TacIPC and DIFFTACTILE lack seamless integration with Isaac Sim's parallel execution model. TacEx provides the integration foundation but does not emphasize comprehensive stress extraction and force decomposition. Taccel and UniVTAC focus on scalability and multi-sensor support but do not provide detailed mechanical ground truth from first principles. Our work builds upon TacEx's integration approach while specifically focusing on complete physics-based force computation: we extract full Cauchy stress tensors, decompose them into normal pressure and tangential traction, and provide these mechanical quantities alongside optional optical rendering. This dual capability—accurate mechanics plus visual reference—within a unified Isaac-integrated framework distinguishes our approach.

\subsection{Multi-Sensor Tactile Frameworks}
In the era of foundation models, the robotics community is actively pursuing unified representations across diverse tactile sensors. Frameworks such as UniTouch \cite{yang2024binding}, AnyTouch \cite{feng2025anytouch} and VTV-LLM \cite{xie2026universal} align heterogeneous tactile signals into shared latent spaces. However, training these unified representations requires massive amounts of paired data across different sensors interacting with identical objects—a process that is prohibitively expensive in the real world.

Our framework addresses this need by providing an extensible sensor interface that enables rapid integration of new sensors with minimal configuration changes. By supporting multiple sensor types (GelSight Mini, DIGIT, 9DTact) within a unified physics-based simulation framework, we enable generation of diverse multi-sensor tactile data for training and evaluation. The modular design allows researchers to add new sensors by simply providing sensor geometry and calibration parameters, facilitating large-scale data generation for multi-sensor representation learning with trustworthy mechanical supervision.

\begin{figure*}[t]
\centering
\includegraphics[width=\textwidth]{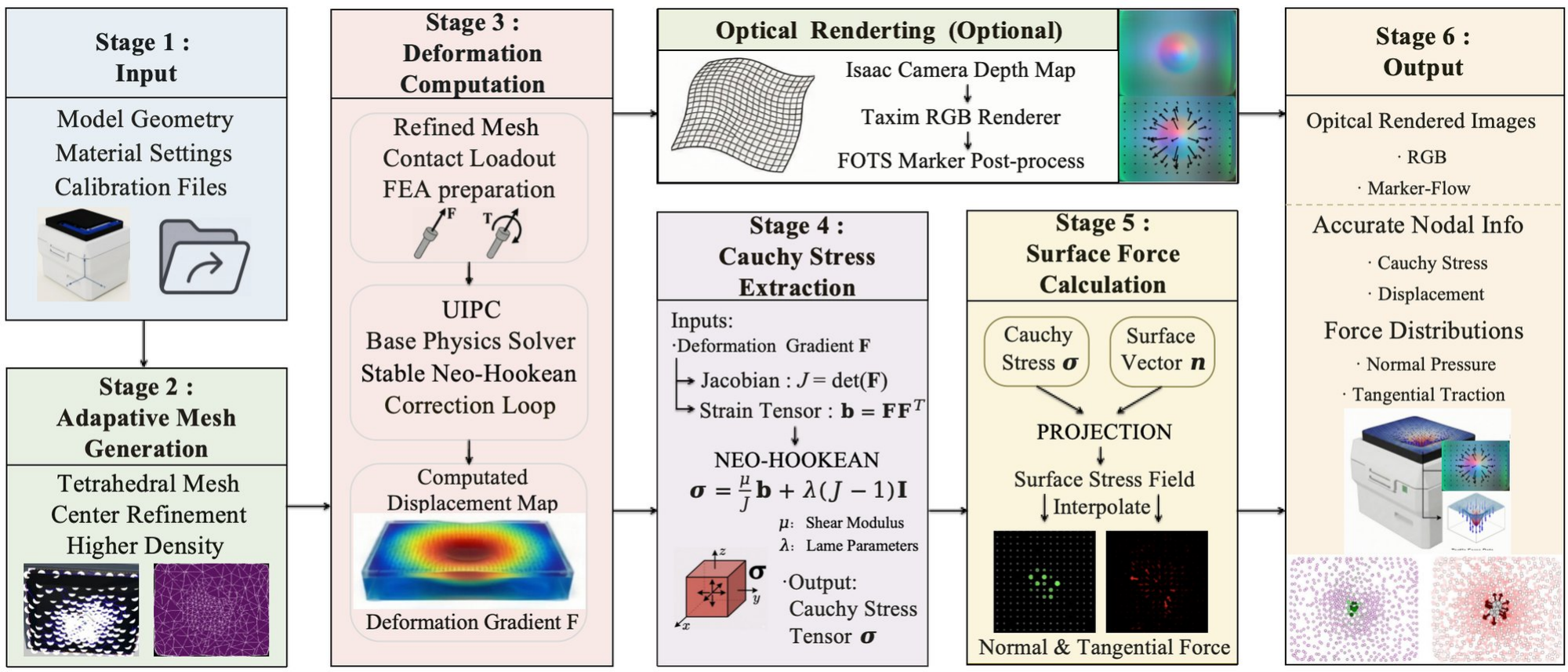}
\caption{TaCauchy complete simulation pipeline. The end-to-end workflow from sensor geometry input to downstream applications, highlighting core components: automatic mesh generation with adaptive refinement, UIPC-FEM solver with direct force computation from hyperelastic constitutive laws, Cauchy stress extraction and traction decomposition, modular sensor interface, optional optical rendering, and integration with Isaac Lab for reinforcement learning. Red borders mark core contribution modules.}
\label{fig:pipeline}
\end{figure*}

\section{Methodology}\label{sec:methodology}

To accurately capture the non-linear deformations and complex stress distributions of heterogeneous visuo-tactile sensors, \textit{TaCauchy} employs a high-fidelity Finite Element Method (FEM) backend powered by the Unified Incremental Potential Contact (UIPC) library. 

\subsection{Mesh Generation and Contact-Surface Refinement}
We model the sensor's elastomer using a tetrahedral mesh generated via the WildMeshing algorithm \cite{hu2020fast}, a robust tetrahedralization framework that guarantees intersection-free, inversion-free, and manifold-valid meshes. The mesh generation pipeline begins with the surface triangulation of the sensor geometry and proceeds to volumetric tetrahedralization subject to user-specified quality constraints. Key parameters include the relative target edge length $l_r$ (controlling global element density), the geometric envelope parameter $\epsilon_r$ (preserving fine surface features), the maximum AMIPS energy threshold (guaranteeing element quality), and the stopping quality $q_{\min}$ (bounding worst-case element aspect ratios).

To improve contact resolution while controlling computational cost, we introduce a geometry-aware adaptive refinement strategy based on WildMeshing's per-vertex \emph{sizing field}. We register a spatial target edge-length function directly into the tetrahedralization kernel. This ensures that element density is modulated during the Delaunay optimization itself, producing well-shaped tetrahedra throughout the transition region.

The sizing field operates as follows. First, the thinnest bounding-box axis is automatically identified as the elastomer pad thickness direction, and vertices lying within 15\% of the thickness from the lower face are labeled as the contact surface. A radial density profile is then applied over that surface: elements within the central region (inner radius $r_c = 40\%$ of the planar half-extent) are assigned a fine target edge length of $l_c \approx 0.8$\,mm, while the periphery retains the global coarse resolution of $l_g \approx 3.2$\,mm. A cosine-smoothed transition band (width $r_t = 20\%$ of the half-extent) bridges the two regions to avoid abrupt stiffness discontinuities. This strategy concentrates computational resources where contact occurs while maintaining overall efficiency. 

\subsection{Hyperelastic Constitutive Model and UIPC Solver}
The soft elastomer pad is modeled using a Stable Neo-Hookean hyperelastic constitutive model. This choice is particularly well-suited for vision-based tactile sensors for several reasons. First, the Neo-Hookean formulation is grounded in the statistical thermodynamics of cross-linked polymer chains \cite{treloar1975physics}, accurately capturing the mechanical behavior of elastomeric materials such as silicone rubber commonly used in tactile sensing applications \cite{lambeta2020digit,lin20249dtact}. Second, the Stable Neo-Hookean variant \cite{smith2018stable} provides superior numerical robustness under extreme deformations and element inversions—conditions frequently encountered during high-frequency contact interactions in manipulation tasks. Third, this model exhibits excellent volume preservation characteristics, which is critical for accurately simulating the incompressible behavior of soft gels (Poisson's ratio $\nu \approx 0.5$) while maintaining computational stability.

Given the Young's modulus $E$ and Poisson's ratio $\nu$ (where $\nu < 0.5$), the Lamé parameters are derived through the standard elasticity relations:
\begin{equation}\label{eq:lame_parameters}
\mu = \frac{E}{2(1+\nu)}, \quad \lambda = \frac{E\nu}{(1+\nu)(1-2\nu)}
\end{equation}
where $\mu$ represents the shear modulus and $\lambda$ governs the volumetric response. The strain energy density function yields the First Piola-Kirchhoff stress tensor $\mathbf{P}$:
\begin{equation}\label{eq:pk_stress}
\mathbf{P} = \mu \mathbf{F} + \lambda(\det(\mathbf{F}) - 1)\det(\mathbf{F})\mathbf{F}^{-T}
\end{equation}
where $\mathbf{F}$ denotes the deformation gradient tensor. 

To robustly resolve dynamic soft-body interactions without mesh inversion or intersection, we utilize the UIPC engine~\cite{li2020incremental}'s optimization-based integration scheme. In our configuration, the Incremental Potential Contact (IPC) barrier distance threshold ($\hat{d}$) is strictly set to 0.001 m, coupled with a velocity threshold ($\epsilon_{v}$) of 0.01 m/s and a frictional contact ratio of 0.5. The Newton solver is configured to permit up to 1024 maximum iterations per time step, ensuring numerical stability even under extreme shear loads.

\subsection{Cauchy Stress Extraction and Traction Decomposition}

A core engineering contribution of our methodology is the direct extraction and decomposition of the Cauchy stress tensor ($\boldsymbol{\sigma}$) from the UIPC deformation field. Unlike learning-based inverse methods that estimate forces from observed deformations or simplified contact models that use empirical spring-damper approximations, our approach computes forces through direct physics-based calculation from first principles.

\subsubsection{Stress Tensor Computation}

The Cauchy stress tensor describes the true stress state in the current (deformed) configuration. For the Stable Neo-Hookean model, the relationship between the First Piola-Kirchhoff stress $\mathbf{P}$ and the Cauchy stress is given by:
\begin{equation}\label{eq:pk_to_cauchy}
\boldsymbol{\sigma} = \frac{1}{J} \mathbf{P} \mathbf{F}^T
\end{equation}
where $J = \det(\mathbf{F})$ is the Jacobian determinant representing volumetric deformation. Substituting Eq.~\eqref{eq:pk_stress} into Eq.~\eqref{eq:pk_to_cauchy} and simplifying yields:
\begin{equation}\label{eq:cauchy_stress}
\boldsymbol{\sigma} = \frac{\mu}{J} \mathbf{b} + \lambda(J - 1) \mathbf{I}
\end{equation}
where $\mathbf{b} = \mathbf{F}\mathbf{F}^T$ is the left Cauchy-Green deformation tensor and $\mathbf{I}$ is the identity tensor. This formulation is grounded in the statistical thermodynamics of cross-linked polymer chains \cite{treloar1975physics}, providing a physically rigorous foundation rather than empirical fitting.

The symmetric Cauchy stress tensor contains nine components (six independent due to symmetry):
\begin{equation}
\boldsymbol{\sigma} = \begin{bmatrix}
\sigma_{xx} & \sigma_{xy} & \sigma_{xz} \\
\sigma_{yx} & \sigma_{yy} & \sigma_{yz} \\
\sigma_{zx} & \sigma_{zy} & \sigma_{zz}
\end{bmatrix}
\end{equation}
where diagonal elements $\sigma_{ii}$ represent normal stresses (tension/compression) and off-diagonal elements $\sigma_{ij}$ represent shear stresses.

\subsubsection{Surface Stress Projection Algorithm}

Our stress extraction module implements a systematic algorithm to project volumetric stress fields onto sensor surfaces. The complete procedure is outlined in Algorithm~\ref{alg:stress_extraction}.

\begin{algorithm}[t]
\caption{Surface Cauchy Stress Extraction}
\label{alg:stress_extraction}
\begin{algorithmic}[1]
\REQUIRE Tetrahedral mesh $(V, T)$, surface triangles $\mathcal{T}_s$, stress field $\{\boldsymbol{\sigma}_i\}$
\ENSURE Surface traction $\{\mathbf{t}_j\}$, normal pressure $\{p_{n,j}\}$, tangential traction $\{\mathbf{t}_{\tau,j}\}$

\STATE Extract surface triangle topology and parent tetrahedra mapping
\STATE $\mathcal{T}_s \leftarrow$ triangles with $\texttt{is\_surf} = \texttt{true}$
\STATE $\{\text{parent}_j\} \leftarrow$ parent tetrahedral IDs for each $j \in \mathcal{T}_s$

\FOR{each surface triangle $j \in \mathcal{T}_s$}
    \STATE Retrieve vertices: $\mathbf{v}_0, \mathbf{v}_1, \mathbf{v}_2 \leftarrow V[\mathcal{T}_s[j]]$
    
    \STATE \textbf{Compute surface normal:}
    \STATE $\mathbf{e}_1 \leftarrow \mathbf{v}_1 - \mathbf{v}_0$, $\mathbf{e}_2 \leftarrow \mathbf{v}_2 - \mathbf{v}_0$
    \STATE $\mathbf{n}_j \leftarrow \frac{\mathbf{e}_1 \times \mathbf{e}_2}{\|\mathbf{e}_1 \times \mathbf{e}_2\|}$
    
    \STATE \textbf{Retrieve parent stress tensor:}
    \STATE $\boldsymbol{\sigma}_j \leftarrow \boldsymbol{\sigma}_{\text{parent}_j}$
    
    \STATE \textbf{Apply Cauchy stress theorem:}
    \STATE $\mathbf{t}_j \leftarrow \boldsymbol{\sigma}_j \cdot \mathbf{n}_j$ \COMMENT{Traction vector}
    
    \STATE \textbf{Decompose into normal and tangential components:}
    \STATE $p_{n,j} \leftarrow \mathbf{t}_j \cdot \mathbf{n}_j$ \COMMENT{Normal pressure (scalar)}
    \STATE $\mathbf{t}_{\tau,j} \leftarrow \mathbf{t}_j - p_{n,j} \mathbf{n}_j$ \COMMENT{Tangential traction (vector)}
\ENDFOR

\RETURN $\{\mathbf{t}_j\}, \{p_{n,j}\}, \{\mathbf{t}_{\tau,j}\}$
\end{algorithmic}
\end{algorithm}

The key mathematical operation is the Cauchy stress theorem, which relates the stress tensor to surface traction:
\begin{equation}\label{eq:cauchy_theorem}
\mathbf{t} = \boldsymbol{\sigma} \cdot \mathbf{n} = \sum_{j=1}^{3} \sigma_{ij} n_j \mathbf{e}_i
\end{equation}
where $\mathbf{e}_i$ are the Cartesian basis vectors. In matrix form:
\begin{equation}
\begin{bmatrix} t_x \\ t_y \\ t_z \end{bmatrix} = 
\begin{bmatrix}
\sigma_{xx} & \sigma_{xy} & \sigma_{xz} \\
\sigma_{yx} & \sigma_{yy} & \sigma_{yz} \\
\sigma_{zx} & \sigma_{zy} & \sigma_{zz}
\end{bmatrix}
\begin{bmatrix} n_x \\ n_y \\ n_z \end{bmatrix}
\end{equation}

\subsubsection{Normal-Tangential Decomposition}

To provide comprehensive mechanical information necessary for 6D force estimation, we mathematically decompose the traction vector into orthogonal components. The normal pressure (scalar) is computed as:
\begin{equation}\label{eq:normal_pressure}
p_n = \mathbf{t} \cdot \mathbf{n} = \mathbf{n}^T \boldsymbol{\sigma} \mathbf{n}
\end{equation}
This quadratic form represents the stress component perpendicular to the surface, where $p_n > 0$ indicates tensile stress and $p_n < 0$ indicates compressive stress. The tangential traction vector, representing shear forces parallel to the surface, is obtained by subtracting the normal component:
\begin{equation}\label{eq:traction_decomp}
\mathbf{t}_{\tau} = \mathbf{t} - p_n \mathbf{n} = \mathbf{t} - (\mathbf{t} \cdot \mathbf{n}) \mathbf{n}
\end{equation}
This decomposition satisfies the orthogonality condition $\mathbf{t}_{\tau} \cdot \mathbf{n} = 0$ by construction, ensuring that $\mathbf{t}_{\tau}$ lies entirely in the tangent plane.

The magnitude of tangential traction $\|\mathbf{t}_{\tau}\|$ quantifies the shear stress intensity, which is critical for analyzing frictional contact and slip conditions. This rigorous decoupling provides pure physical ground-truth force information required for evaluating interactions across heterogeneous sensors entirely within simulation, enabling downstream learning algorithms to leverage reliable mechanical supervision without the uncertainty introduced by inverse estimation or the inaccuracy of simplified empirical models.

Algorithm~\ref{alg:simulation_loop} summarizes the complete simulation pipeline, integrating mesh generation, UIPC-FEM dynamics, stress extraction, and force decomposition into a unified computational framework.

\begin{algorithm}[h]
\caption{TaCauchy Simulation Loop}
\label{alg:simulation_loop}
\begin{algorithmic}[1]
\REQUIRE Sensor geometry $\mathcal{G}$, material $(E, \nu)$, time step $\Delta t$
\ENSURE Stress $\mathcal{S}_t$ and force $\mathcal{F}_t$ at each step $t$

\STATE Generate tetrahedral mesh $(V, T)$ via WildMeshing with contact-surface adaptive sizing field $s(\mathbf{x})$
\STATE Initialize UIPC solver and force extractors

\FOR{$t = 0$ to $T_{\text{max}}$}
    \STATE \textbf{Physics:} $\mathbf{x}_{t+1} \leftarrow$ UIPC.step($\mathbf{x}_t$, $\Delta t$)
    
    \STATE \textbf{Stress:} Compute $\boldsymbol{\sigma}_i = \frac{\mu}{J_i} \mathbf{b}_i + \lambda(J_i - 1) \mathbf{I}$ for each tet
    
    \FOR{each surface triangle $j$}
        \STATE Compute normal: $\mathbf{n}_j$, traction: $\mathbf{t}_j = \boldsymbol{\sigma}_i \cdot \mathbf{n}_j$
        \STATE Decompose: $p_{n,j} = \mathbf{t}_j \cdot \mathbf{n}_j$, $\mathbf{t}_{\tau,j} = \mathbf{t}_j - p_{n,j} \mathbf{n}_j$
    \ENDFOR
    
    \STATE \textbf{Contact:} Get gradient $\mathbf{g}_k$, compute $\mathbf{f}_k = -\mathbf{g}_k/\Delta t^2$
    
    \FOR{each contact vertex $k$}
        \STATE Compute normal $\mathbf{n}_k$, area $A_k$
        \STATE Decompose: $\mathbf{f}_{n,k} = (\mathbf{f}_k \cdot \mathbf{n}_k) \mathbf{n}_k$, $\mathbf{f}_{\tau,k} = \mathbf{f}_k - \mathbf{f}_{n,k}$
        \STATE Pressure: $p_k = \|\mathbf{f}_k\|/A_k$
    \ENDFOR
\ENDFOR
\end{algorithmic}
\end{algorithm}

\subsection{Physically-Grounded Optical Rendering}

To provide high-fidelity visual feedback consistent with mechanical ground truth, our framework integrates a hybrid optical simulation module. Unlike existing decoupled approaches where optical rendering and mechanical physics operate in isolation, our rendering process is physically constrained by the FEM simulation.

In traditional tactile simulators such as Taxim \cite{si2022taxim}, the sensor's elastomer pad often exhibits unnatural interpenetration with objects due to the lack of physical contact constraints in the optical depth generation. In the \textit{TaCauchy} framework, we resolve this discrepancy by leveraging the non-penetrating contact interface computed by our FEM solver. The deformed surface mesh of the elastomer pad acts as a strict physical boundary for the optical module; specifically, the depth maps used for rendering are constrained to the volume defined by the FEM-calculated mesh deformation. This ensures that the visual signal accurately reflects the physical contact state, significantly enhancing the credibility and realism of the generated tactile images.For sensors featuring embedded tactile markers, we incorporate the FOTS \cite{zhao2024fots} framework to simulate marker dynamics. Through this integration, our framework establishes a faithful mapping from mechanical deformation to optical response, ensuring that the synthesized tactile data is both physically grounded and visually authentic.

\section{Experiments}\label{sec:experiments}

\subsection{Simulation Experimental Setup}

All experiments were conducted on a workstation with an AMD Ryzen 9 9950X processor and RTX 5090 GPU, using Isaac Sim 5.1.0~\cite{isaacsim2023}, Isaac Lab 0.51.1~\cite{mittal2023orbit}, and UIPC for soft-body dynamics. Three tactile sensors were evaluated (GelSight Mini, DIGIT, 9DTact), each mounted on a Franka Panda manipulator.

Taking GelSight Mini as reference, the refined elastomer pad mesh contains 1,920 tetrahedral elements with 553 vertices, 358 surface vertices, and 712 triangular faces. The adaptive refinement achieves 2.90$\times$ increase in tetrahedral count (from 661 baseline to 1,920 refined elements) while reducing contact-center edge length from 1.82 mm to 0.68 mm, yielding 7.35$\times$ higher mesh density in the critical contact zone. Performance benchmarks show 33.40 FPS (headless) and 9.23 FPS (GUI) for single environments. Scaling to 60 parallel environments achieves 9.25 FPS (headless) and 7.59 FPS (GUI), corresponding to 555 FPS and 455 FPS aggregate throughput respectively. Physics dominates computation (88.92 ms headless, 102.72 ms GUI) while stress extraction remains minimal ($<$1 ms). This computational efficiency enables rapid large-scale contact data generation for tactile perception research.

\subsection{Physical Validation Experiment}

\begin{figure}[t]
\centering
\begin{tabular}{@{}c@{\hspace{0.02\columnwidth}}c@{}}
\includegraphics[width=0.49\columnwidth]{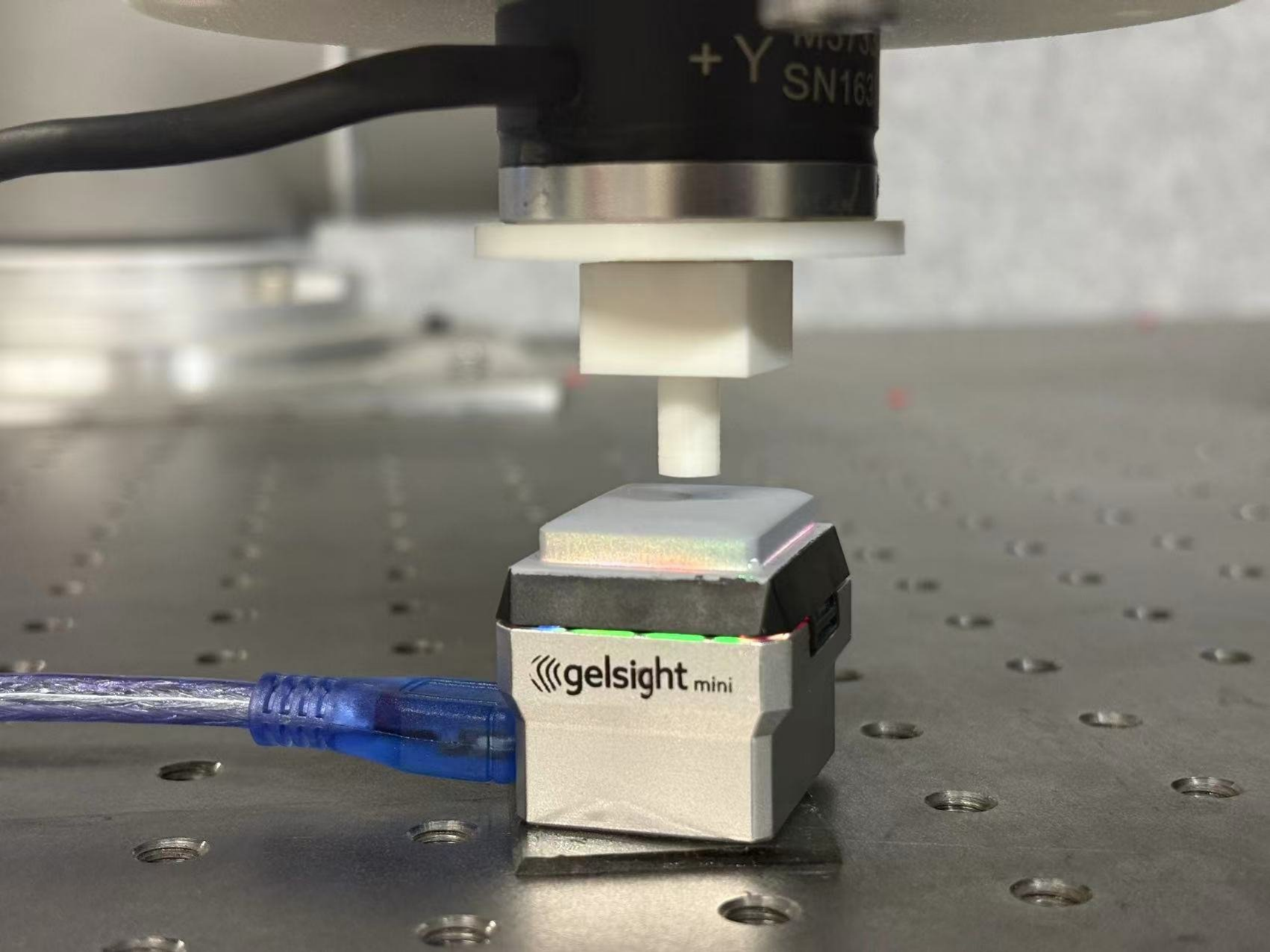} &
\includegraphics[width=0.49\columnwidth]{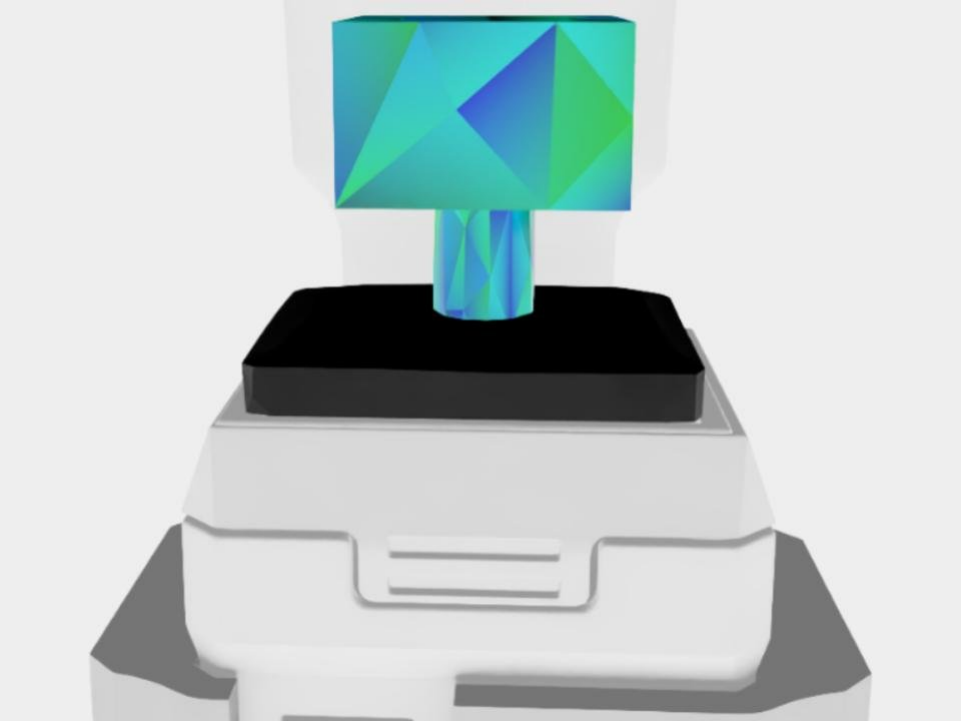} \\
(a) Real Setup & (b) Simulation Environment
\end{tabular}
\caption{Experimental setup comparison: (a) Physical setup with Universal Robots UR5 arm, M3733C force sensor, and GelSight Mini tactile sensor; (b) Corresponding simulation environment in Isaac Sim with identical geometric configuration.}
\label{fig:exp_setup}
\end{figure}

To validate the accuracy of our physics-based simulation framework, we conducted a series of controlled contact experiments comparing real-world tactile sensor responses with simulated outputs under identical loading conditions. The experimental setup consisted of a Universal Robots UR5 collaborative robot arm equipped with a 6-axis force/torque sensor (M3733C) mounted at the end-effector. A cylindrical indenter (diameter 5 mm, height 10 mm) was rigidly attached to the force sensor, positioned vertically downward to contact a GelSight Mini sensor fixed on the experimental platform. Figure~\ref{fig:exp_setup} shows the real experimental setup and its corresponding simulation environment.

The robot was programmed to perform controlled vertical pressing motions, gradually increasing the contact force while maintaining alignment perpendicular to the sensor surface. During each trial, the force sensor recorded the vertical force component $F_z$ at 100 Hz sampling rate, providing ground truth force measurements with 0.0001 N resolution. Simultaneously, the GelSight Mini sensor captured tactile images at 25 FPS, synchronized with the force readings through hardware triggering. Six experimental trials were conducted with target forces ranging from 1.2556 N to 4.7332 N, covering the typical operating range of vision-based tactile sensors in manipulation tasks.

For each experimental trial, we replicated the contact scenario in our Isaac Sim simulation environment using identical geometric models and material parameters. The cylindrical indenter was modeled as a rigid body with the same dimensions as the physical object, while the GelSight Mini sensor utilized the finite element mesh and material properties described in Section~\ref{sec:experiments}. The simulation applied vertical displacement boundary conditions to the indenter until the computed contact force matched the experimentally measured $F_z$ value within 0.0800 N tolerance. At force equilibrium, the simulated tactile image was rendered using the same photometric model employed for real-time sensor operation.

\begin{figure}[t]
\centering
\includegraphics[width=\columnwidth]{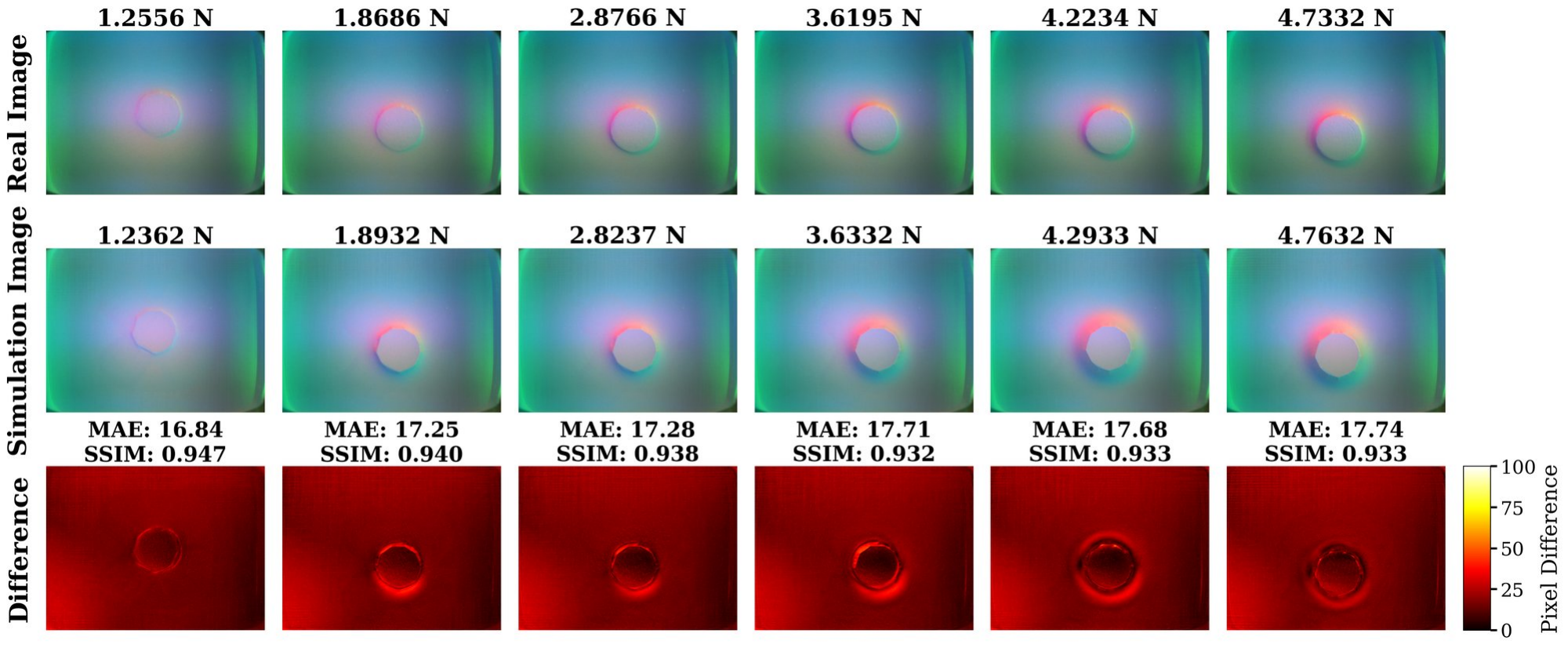}
\caption{Comparison of real and simulated tactile images across six force levels. The first row shows real sensor images, the second row shows simulated images, and the third row shows pixel-wise differences. Each column corresponds to a trial with increasing vertical force $F_z$, demonstrating consistent deformation patterns between physical experiments and simulation.}
\label{fig:trial_comparisons}
\end{figure}

Fig.~\ref{fig:trial_comparisons} presents side-by-side comparisons of real and simulated tactile images for all six force levels. Visual inspection reveals strong qualitative agreement in contact geometry, deformation patterns, and intensity distributions. The contact area expands progressively with increasing force, exhibiting circular symmetry consistent with the cylindrical indenter geometry. Both real and simulated images capture the characteristic bright center region surrounded by darker peripheral zones.

To quantify similarity, we computed four metrics: SSIM, NCC, Histogram Correlation, and PSNR. SSIM evaluates structural similarity through luminance, contrast, and structure. NCC measures normalized correlation between intensities. Histogram Correlation assesses global intensity distribution, while PSNR quantifies pixel-wise fidelity.

\begin{figure}[t]
\centering
\includegraphics[width=\columnwidth]{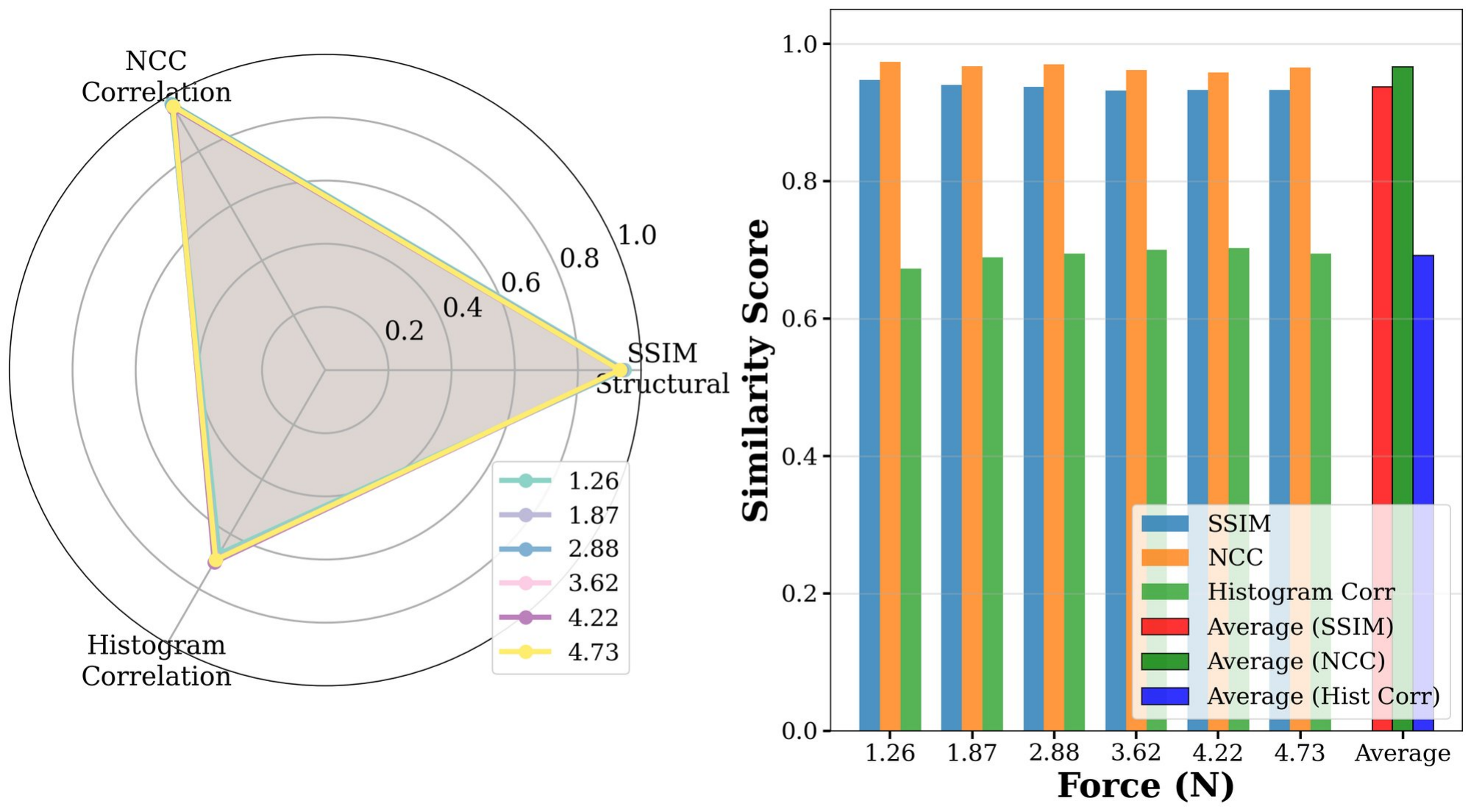}
\caption{Quantitative comparison metrics across six experimental trials. Left: Radar chart showing SSIM, NCC, and Histogram Correlation for each force level. Right: Bar chart comparing the three similarity metrics across all trials with average values. Force values are displayed with two decimal places for clarity.}
\label{fig:metrics_comparison}
\end{figure}

\begin{figure}[t]
\centering
\begin{tabular}{@{}c@{\hspace{0.02\columnwidth}}c@{}}
\includegraphics[width=0.49\columnwidth]{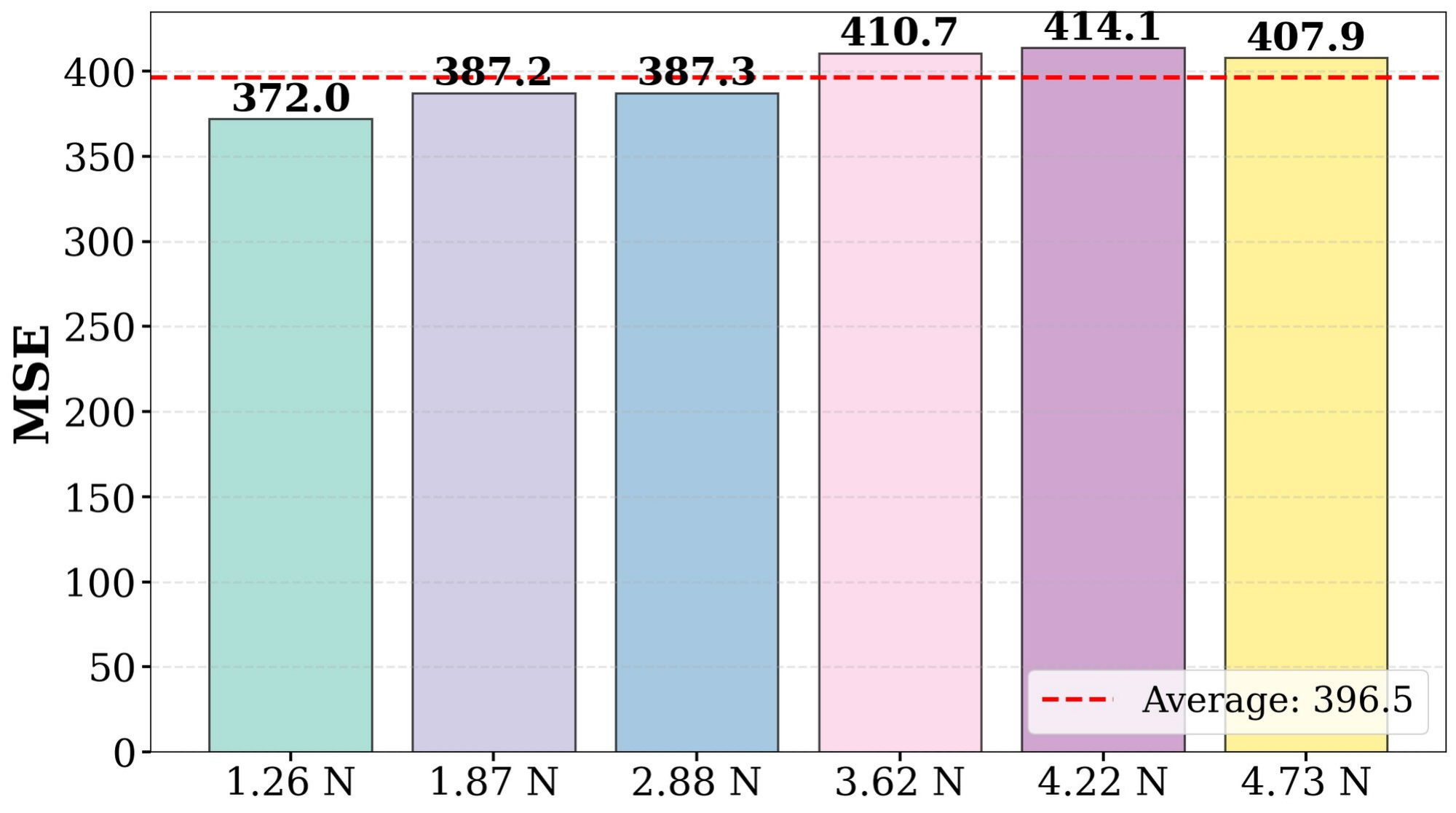} &
\includegraphics[width=0.49\columnwidth]{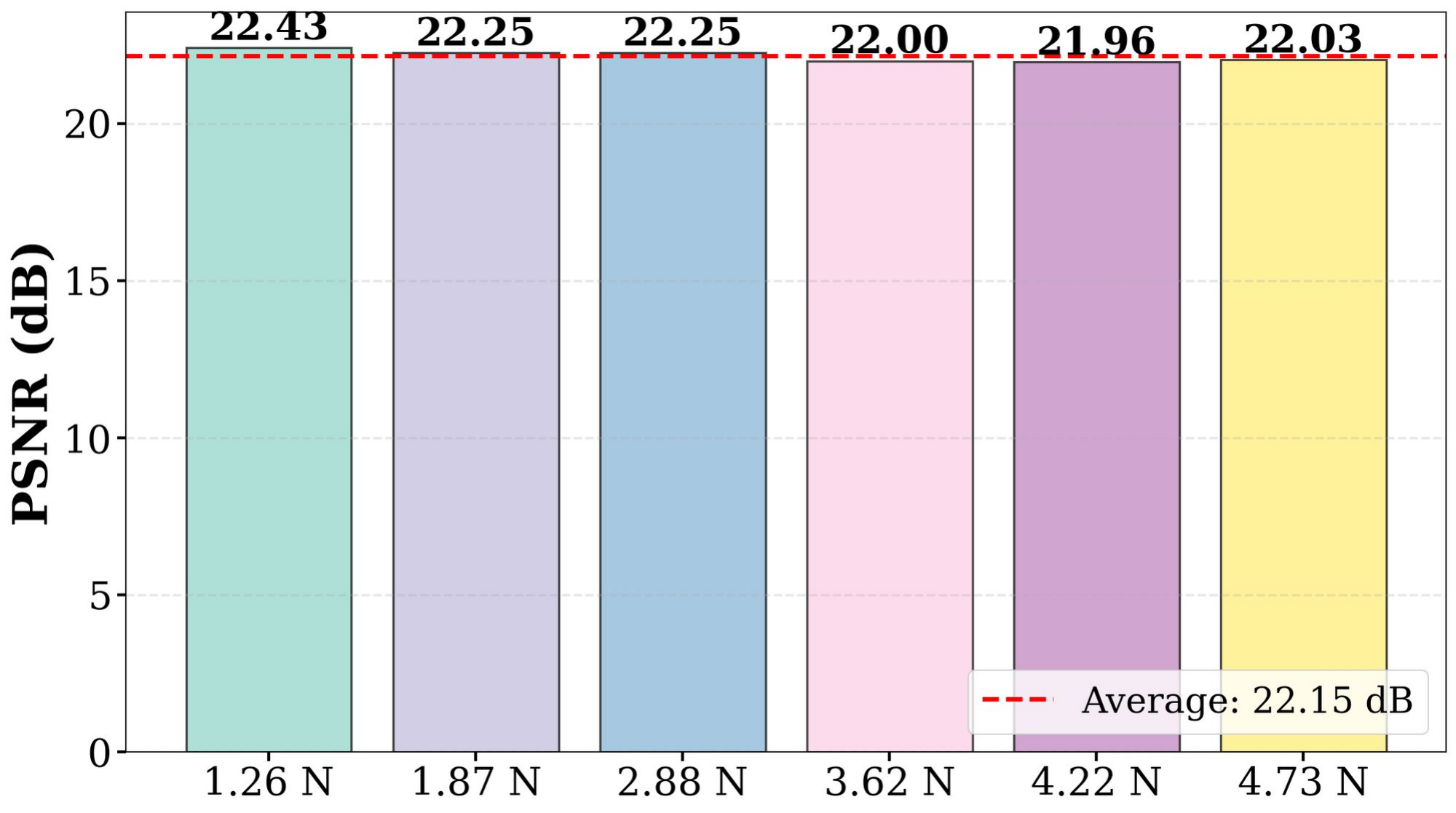} \\
(a) & (b)
\end{tabular}
\caption{Pixel-wise error metrics across six force levels. (a) Mean Squared Error (MSE) showing consistently low pixel-wise discrepancies with average line. (b) Peak Signal-to-Noise Ratio (PSNR) demonstrating acceptable image fidelity across all trials with average line. Force values are displayed with two decimal places for clarity.}
\label{fig:mse_psnr_comparison}
\end{figure}

Fig.~\ref{fig:metrics_comparison} summarizes the quantitative evaluation. SSIM scores consistently exceed 0.93 (average 0.9377), confirming strong structural similarity. NCC values average 0.9669, indicating accurate spatial pattern capture despite lighting variations. Histogram Correlation achieves 0.6920, reflecting reasonable intensity agreement with higher sensitivity to lighting differences. PSNR averages 22.13~dB (Fig.~\ref{fig:mse_psnr_comparison}b), representing acceptable fidelity. MSE values (Fig.~\ref{fig:mse_psnr_comparison}a) range from 372.0 to 414.1.

As contact force increases from 1.2556~N to 4.7332~N, metrics exhibit slight degradation: SSIM decreases from 0.947 to 0.933, NCC drops from 0.974 to 0.965, and PSNR declines from 22.43~dB to 22.03~dB. This correlates with increasing gel deformation magnitude, which introduces greater geometric nonlinearity and sensitivity to material uncertainties. Nevertheless, all metrics remain within acceptable ranges throughout the force spectrum, demonstrating framework robustness under larger-deformation scenarios.

The experimental validation demonstrates that our physics-based simulation framework accurately reproduces the mechanical behavior of vision-based tactile sensors under controlled loading conditions. The high SSIM scores indicate that the simulation captures not only the overall deformation magnitude but also the fine-scale spatial patterns critical for downstream perception tasks. Minor discrepancies between real and simulated images can be attributed to several factors: (1) slight misalignments in the relative pose between the sensor and the object compared to the physical setup, (2) inherent discrepancies between the finite element discretization and the continuous deformation of the physical elastomer, (3) material parameter uncertainties in the gel elastomer's hyperelastic properties, and (4) lighting fluctuations and color calibration variations in the real sensor hardware. Despite these limitations, the quantitative metrics confirm that our approach provides sufficient fidelity for training and evaluating vision-based tactile manipulation policies in simulation, ensuring the structural accuracy necessary for effective Sim-to-Real transfer.

\subsection{Contact Mode Analysis in Simulation}

\begin{figure}[t]
\centering
\includegraphics[width=0.75\columnwidth]{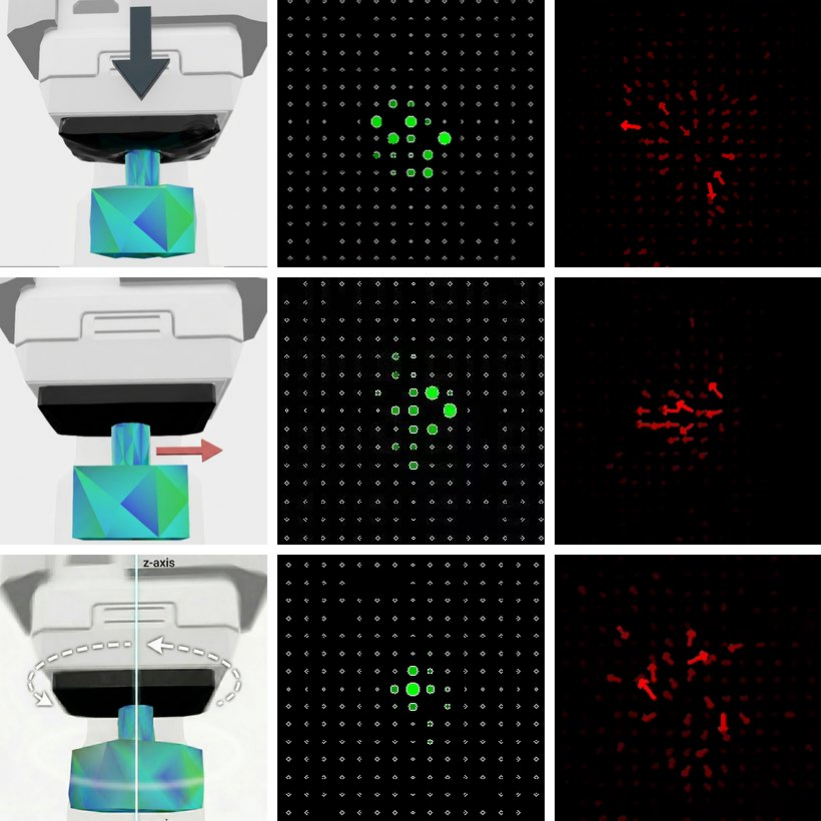}
\caption{Comprehensive visualization of three fundamental contact modes in simulation. Each row represents a different contact mode: (top) normal pressing, (middle) tangential translation, and (bottom) rotational motion. Columns show: (left) geometric configuration and contact interaction view, (center) normal force field distribution, and (right) tangential force field distribution. The force field visualizations reveal the underlying mechanical interactions that generate the observed tactile deformation patterns. Force magnitudes are displayed with two decimal places for precision.}
\label{fig:contact_modes_combined}
\end{figure}

To evaluate the mechanical fidelity of \textit{TaCauchy}, we conducted a qualitative analysis of three canonical contact modes using a cylindrical indenter interacting with the GelSight Mini elastomer: \textbf{normal pressing}, \textbf{lateral translation}, and \textbf{axial rotation}. As visualized in Fig.~\ref{fig:contact_modes_combined}, the framework successfully resolves the distinct force field signatures associated with each interaction.

In \textbf{normal pressing}, the framework captures a symmetric normal pressure distribution concentrated at the contact manifold, while simultaneously resolving a radially symmetric tangential traction field expanding from center to periphery. This accurately reflects the material's volume-preservation and Poisson's effect, where vertical displacement induces lateral surface expansion.

During \textbf{lateral translation} (sliding), \textit{TaCauchy} resolves the characteristic symmetry breaking in the contact stress field. The normal force exhibits a pronounced pressure gradient shifted toward the leading edge—a signature of elastomer accumulation in the direction of travel. The tangential traction vectors are oriented uniformly opposite to the sliding direction, representing global frictional resistance. Notably, at the leading edge where compression occurs, tangential forces display a radial spreading tendency, reflecting local material flow and stress redistribution.

In the \textbf{axial rotation} (torsion) case, the normal force maintains symmetric distribution while the framework extracts the complex circumferential traction field. \textit{TaCauchy} resolves a vortex-like vector field where tangential forces are precisely oriented tangentially to the local rotation path, demonstrating the ability to decouple multi-axial stress components.

Experiments prove our framework accurately simulates rigid-soft interactions. Its force visualizations reveal underlying physics, providing essential training data for learning-based tactile perception and robust manipulation policies where force-visual relationships are key.

\subsection{Multi-Sensor Force Field Validation}
To evaluate framework extensibility across diverse sensor platforms, we conducted force field analysis using three mainstream tactile sensors: GelSight Mini, DIGIT, and 9DTact. Adaptive FEM meshes were generated for each sensor geometry. For 9DTact, we modeled its dual-layer structure with a soft translucent base layer and stiffer black surface layer, ensuring accurate representation of heterogeneous material properties.

\begin{figure}[t]
\centering
\includegraphics[width=\columnwidth]{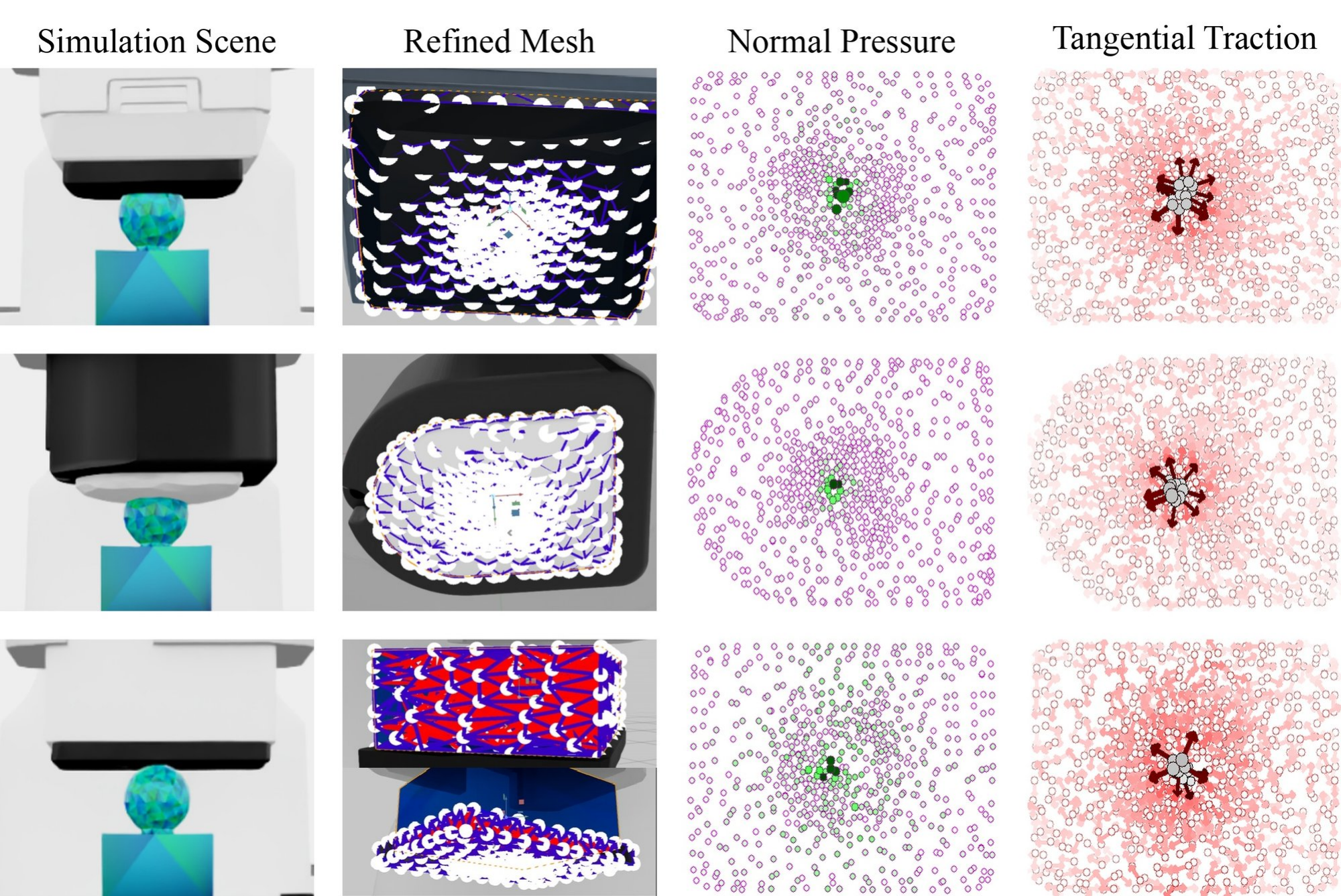}
\caption{Multi-sensor force field validation with spherical indenter contact. Rows: (top) GelSight Mini, (middle) DIGIT, (bottom) 9DTact. Columns: simulation scene, adaptive FEM mesh, normal pressure distribution, tangential traction field. Force magnitudes are displayed with two decimal places for precision.}
\label{fig:multi_sensor_meshes}
\end{figure}

We simulated spherical indenter contact against each sensor's elastomer surface. Fig.~\ref{fig:multi_sensor_meshes} presents the results: each row corresponds to one sensor (top: GelSight Mini, middle: DIGIT, bottom: 9DTact), with columns showing (left to right) simulation scene, adaptive mesh, normal pressure distribution, and tangential traction field. Normal pressure exhibits concentrated compressive stress at contact center with radial decay, while tangential traction displays radial spreading from Poisson's effect. Results demonstrate accurate sensor-specific mechanical responses across diverse geometries.

For sensors based on internal illumination principles, such as GelSight and DIGIT, we applied physically-constrained optical rendering to synthesize tactile images. To validate the realism of the generated signals, we selected five test objects with distinct geometric features—ranging from sharp edges to smooth curvatures. Through this integration, our TaCauchy framework establishes a faithful mapping from mechanical deformation to optical response, ensuring that the synthesized tactile data is both physically grounded and visually authentic.

\begin{figure}[t]
\centering
\includegraphics[width=\columnwidth]{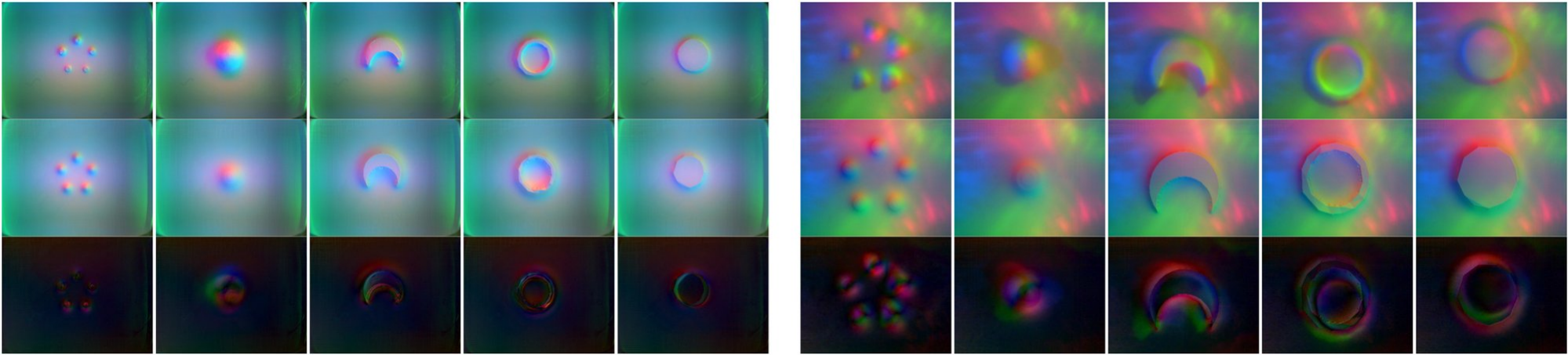}
\caption{Visual comparison between real and simulated tactile images across multiple sensor platforms. The figure displays side-by-side comparisons for GelSight Mini (left) and DIGIT (right) sensors. Each column shows: real tactile data (top row), TaCauchy-simulated results (middle row), and pixel-wise difference maps (bottom row).}

\label{fig:multi_sensor_comparison}
\end{figure}

Fig.~\ref{fig:multi_sensor_comparison} presents a comprehensive visual comparison between real-world captures and TaCauchy-simulated outputs across multiple sensor platforms. Quantitative evaluation demonstrates strong agreement: GelSight Mini achieves an average SSIM of 0.9277 and PSNR of 22.45~dB, while DIGIT achieves an average SSIM of 0.9215 and PSNR of 25.42~dB. The consistently high SSIM values ($>0.92$) indicate excellent structural similarity in deformation patterns, while the low-magnitude difference maps in the bottom row confirm high visual fidelity. Most minor discrepancies are concentrated in boundary regions where lighting conditions and material imperfections in physical sensors introduce subtle variations. Notably, the framework accurately reproduces the distinct marker patterns of GelSight Mini and the characteristic appearance of DIGIT, validating its capability to generalize across diverse vision-based tactile technologies.

\section{Conclusion}\label{sec:conclusion}

We presented \textit{TaCauchy}, an extensible FEM framework that integrates physics-based force computation into Isaac Sim for vision-based tactile sensor simulation. By directly computing Cauchy stress tensors from hyperelastic constitutive laws, our framework provides trustworthy mechanical ground truth without empirical estimation uncertainties. The modular design enables rapid integration of diverse sensors (GelSight Mini, DIGIT, 9DTact), while physical validation confirms high fidelity with SSIM $> 0.92$. \textit{TaCauchy} establishes a robust platform for downstream robotic manipulation and multi-sensor representation learning.

Despite its capabilities, \textit{TaCauchy} faces limitations regarding material calibration, as physical elastomers exhibit time-varying properties (e.g., wear and hysteresis) that are difficult to model continuously. Additionally, the computational overhead of high-resolution FEM currently constrains the massive parallelization typically desired for ultra-fast policy learning. Future work will focus on deploying this framework within large-scale Deep Reinforcement Learning (DRL) pipelines for contact-rich manipulation tasks. By leveraging our comprehensive mechanical ground truth, we aim to train robust, force-aware control policies capable of seamless zero-shot sim-to-real transfer.

\bibliographystyle{IEEEtran}
\bibliography{references}

\end{document}